# Comparison of Scanned Administrative Document Images

Elena Andreeva[1], Vladimir V. Arlazarov[1,2], Oleg Slavin[2,3], Aleksey Mishev[4]
[1]Smart Engines Service LLC, Moscow, Russia
[2]Federal Research Center "Computer Science and Control" of Russian Academy of Sciences, Moscow, Russia
[3]Moscow Institute of Physics and Technology, Moscow, Russia
[4]National Research Nuclear University MEPhI (Moscow Engineering Physics Institute), Moscow, Russia

## ABSTRACT

In this work the methods of comparison of digitized copies of administrative documents were considered. This problem arises, for example, when comparing two copies of documents signed by two parties in order to find possible modifications made by one party, in the banking sector at the conclusion of contracts in paper form. The proposed method of document image comparison is based on a combination of several ways of image comparison of words that are descriptors of text feature points. Testing was conducted on public Payslip Dataset (French). The results showed the high quality and the reliability of finding differences in two images that are versions of the same document.

## 1. INTRODUCTION

With the spread of electronic document automation [1], the need to work with both electronic versions of documents and digitized copies of paper documents increases. Comparison of two document versions is an essential part of the document processing. The task of document comparison arises, for example, when signing contracts between the two parties, when these processes occur at a different time. The client can print out and sign the contract outside the bank, consequently, he can make changes to the contract for his advantage.

The paper describes a method that allows finding modifications in digitized (scanned) images of business document pages, while the comparison of the main objects of the document – words – is based on a combination of comparison methods using character recognition and adaptive pixel-by-pixel comparison. Not only scanned images of documents can be considered but also images that were obtained using mobile camera if the document was localized in the image, for example, as in [2]. For noisy images, the method of numerical reconstruction [3] by using regularization algorithms can be used.

Existing methods of document comparison are divided into the following groups. The first group of methods is used to compare texts of electronic versions of documents, so such methods are not adapted for comparing images of documents, but can detect not only structural but also semantic modifications [4]. The second group is used for the comparison of document images without character recognition, in particular for layout comparison of documents [5, 6]. Disadvantages of these methods are that only the spatial content of the document is compared [6]. Testing the methods described in [5] was not carried out on public dataset.

The third group is used for document forgery detection. These methods consist of checking whether the test image of the document is forged, having a train set of genuine documents. Such methods [7-12] is not intended to find modifications between the two document images. But the algorithms that were used in those methods [8, 12], for example, the algorithm RAST (Recognition by Adaptive Subdivision of Transformation space) [13], which is used for image matching, can be used for pairwise image comparison.

In the RAST algorithm two images are matched to each other and the corresponding score is calculated. This matching score can be considered as the number of identical characters that have the same position in both document images. The algorithm aligns them to maximize the matching score. Alignment is the determination of the transformation space with parameters $(t_x, t_y, s, \alpha)$, where $t_x$ and $t_y$ are the shift in $x$ and $y$ directions correspondingly, $s$ - scaling factor, $\alpha$ - rotation angle required for optimal alignment of the two documents. The transformation space is initialized using $[t_{x_{min}}, t_{x_{max}}] \times [t_{y_{min}}, t_{y_{max}}] \times [s_{min}, s_{max}] \times [\alpha_{min}, \alpha_{max}]$ with the initial parameter ranges.

A simple way of comparing two document images is the character recognition of document image and comparison of text representations using special software, for example, Microsoft Office. The disadvantage of this method is the need to consider the inevitable recognition errors and errors in extracting text attributes (sizes, styles, etc.).

In this paper, we suggest the method that allows comparing two document images and locating modifications that were made in one of the documents. Methods that were considered above have disadvantages when applied to such tasks because some of them compare only text representation of documents, some - only layout of the documents. In our method, we used character recognition for comparison of text representation of the documents taking into account errors of recognition and pixel-by-pixel comparison of word images to decrease the number of false positives.

## 2. METHOD DESCRIPTION

Let us consider the problem of finding inconsistencies between two scanned images of documents. One image corresponds to a reference document and the other - to a test document. In the test image, it is necessary to find possible modifications. In this paper, the following types of modifications were considered: replacement of one or more characters in a word, replacing one word with another, adding/removing a word or group of words, adding/removing a text line.

The result of the comparison should be a list of modifications indicating the type of modification, the recognized word and the bounding rectangle. Also the bounding rectangles of modified objects (words) should be selected in the reference and test images.

The purpose of the comparison is to find modifications, while it is necessary to minimize the number of type I errors (false positives) and type II errors (false negatives).

The problem of comparison scanned images of documents was considered in [14], where lines were coordinated using comparison of bounding rectangles. The word coordination was carried out using Levenshtein distance and comparison of words bounding rectangles.

Proposed comparison method is based on the document presentation as a set of text feature points. A text feature point is a word in the document image, which is on the one hand a sequence of alphabet characters, and on the other hand - a part of the document image enclosed in a bounding rectangle. Thus, we obtain that the text feature point can be represented as $W = (Ker(W), Box(W))$, where $Ker(W)$ is a kernel of text feature point, that is a sequence of characters of a word consisting of symbols of certain alphabet, and $Box(W) = (x, y, w, h)$ is a bounding rectangle, where $(x, y)$ is coordinates of the upper left point, $w$ is width and $h$ is height of this rectangle.

The concept of a text feature (key) point [15] will be considered as a sequence of characters of the word and its properties. The term text feature point corresponds to the generally accepted concept of a feature (key) point consisting of a pair of coordinates and a descriptor that allows distinguishing a feature point from adjacent points of the image. Known examples of descriptors are SIFT, SURF, ORG, YAP descriptors [16]. In the case of a text feature point, the descriptor is the kernel $Ker(W)$, and there may be feature points with the same descriptors, but with different localization in the document. To detect text feature points, various methods for detecting text in an image can be used, for example, [17]. In the case of document images we used the Tesseract OCR system.

The method of comparing images of documents proposed in this paper is based on a comparison of text feature points detected in the image of the reference document with similar text feature points of the test document image. The similarity measure of text feature points will be based in the first case on the Levenshtein distance, and secondly, on the distance between the word images which introduced below.

Document image comparison method consists of the following steps:

1. Image preprocessing

The reference and test images were converted into grayscales and auto-contrasted. Then images were rotated so that the lines of the text in the image were strictly horizontal, for this the Fast Hough Transform algorithm [18] was used.

2. Segmentation of the text in the image - finding the boxes of words.

The first stage of finding the boxes of words in the image is the image segmentation into text lines, the second – the segmentation of the resulting lines into words. Morphological operations of erosion, dilation, opening and closing to a grayscale image were used for segmentation, as well as projections on horizontal and vertical axes. The operation options related to the sizes of characters and distance between lines was determined according to the known properties of the characters from the reference image.

3. Search for modifications based on character recognition and word image analysis.

Let a document be defined by set of lines $D = \{L_i^d\}_{i=1}^{|D|}$, a line - by a set of words $L^d = \{W_j^l\}_{j=1}^{|L_d|}$, and a word - by a text feature point $W^1 = (Ker(W), Box(W))$.

The important stage in document image comparison is the coordination of lines and words, running through the comparison of text feature points constituting a line. Two lines are considered coordinated if the proportion of coordinated words in these lines is greater than the given threshold $line\_simil$.

A combination of several methods is used to compare words. The first method is based on the character recognition of the text obtained by applying appropriate algorithms to the scanned images. The recognition tool was free software Tesseract Open Source OCR Engine (version 4.0 was used) [19].

The main advantages of Tesseract OCR are that it is free software, where recognition results are presented in the HOCR format (HTML OCR), which contains information about the coordinates of the boxes of recognized words. Errors that are specific to the Tesseract OCR, such as wrong page-layout analysis and the errors in character recognition make difficult to compare documents and insert false positives in the comparison results.

Let us consider the comparison of words that are presented in the form of text feature points. Using the Levenshtein distance [20] $lev(Ker(W_1), Ker(W_2))$, a similarity coefficient for the two words is determined as follows:

$$coeff_{OCR}(W_1, W_2) = 1 - lev(Ker(W_1), Ker(W_2)) / \max(|Ker(W_1)|, |Ker(W_2)|), \qquad (1)$$

where $lev(Ker(W_1), Ker(W_2))$ - Levenshtein distance (edit distance).

Two words $W_1$ and $W_2$ are considered coordinated if $coeff_{OCR}(W_1, W_2)$ is greater than the specified threshold $word\_ocr\_simil$.

For comparison of the recognized words, it is necessary to consider the specificity of Tesseract recognition in editorial instructions at calculation of Levenshtein distance:

- ignoring case, as one of the common recognition errors is the change of case for random letters inside the word;
- some characters, such as short and long dashes, hyphens and minus signs, different types of quotes are considered equal as they can be replaced by each other in recognition;
- ignoring punctuation because possible errors are insignificant when comparing two documents;
- identification of characters that are similar in terms of the recognition mechanism, such as the letter "O" and the number "0".

In the work of the described method, false positives associated with recognition errors occur. Their number can be reduced using another method of word comparison based on pixel-by-pixel comparison.

The second method of word comparison is comparing the grayscale images of words enclosed in bounding rectangles. Since $W = (Ker(W), Box(W))$, there are two images $I_m$ and $I_t$, corresponding to the reference and test images of words enclosed in the rectangles $Box(W_m)$ and $Box(W_t)$.

In the work [21] the method of search for the optimal position of binary images for their comparison is offered. When obtaining the image of the document, different effects of digitization arise. From one prototype can be obtained different images that differ from each other and the prototype, also there are distortions due to random noise, inclinations, shifts and others. To take into account the effects of digitization, a special function of closeness is used, the minimization of which when shifting images in the vertical and horizontal directions gives the optimal position of the images.

To compare the word images the modification of the method described in [21] and the RAST method is used.

For each image $I$ an extended image $O^{(1)}(I)$ is constructed, including the image $I$ itself and its unit neighborhood. The distance between the images $I_m$ and $I_t$ is calculated as:

$$dist(I_m, I_t) = \sum_{i=1}^{s}\sum_{j=1}^{n}(\max(0, t_{ij} - O_{m,ij}^{(1)}) + \max(0, m_{ij} - O_{t,ij}^{(1)})), \qquad (2)$$

where $O^{(1)}(I_m) = \| O_{m,ij}^{(1)} \|$ and $O^{(1)}(I_t) = \| O_{t,ij}^{(1)} \|$ are extended images of $I_m$ and $I_t$.

When comparing images $I_m$ and $I_t$, several shifts of the image $I_t = \| t_{ij} \|$ in different directions, as well as rotation at different angles were carried out. The distance between centered images $I_m$ and $I_t$ is the minimum value of the obtained values

$$dist(I_m, I_t) = \min(dist(I_m, I_t^{(x,y,\alpha)})), \qquad (3)$$

where $(x, y, \alpha)$ is the value of the shift along the $x$ and $y$ correspondingly, and $\alpha$ is rotation angle, and

$$(x, y, \alpha) \in [x_{min}, x_{max}] \times [y_{min}, y_{max}] \times [\alpha_{min}, \alpha_{max}]. \qquad (4)$$

After the distance between two images of words has been determined, it is necessary to introduce the concept of word similarity coefficient based on this distance:

$$coeff_{pix}(W_m, W_t) = \frac{dist(I_m, I_t)}{\max(sum(I_m), sum(I_t))}, \qquad (5)$$

where $sum(I)$ is the sum of the pixels in the image that actually shows the number of "black pixels" that are part of the symbol.

Two words $W_m$ and $W_t$ are considered equal if the word similarity coefficient $coeff_{pix}(W_m, W_t)$ is less than the $word\_pixel\_coeff$ threshold. An example of finding the best position and word overlapping is shown in Fig. 1.

| Image $I_m$ | Image $I_t$ | Overlapping |
|---|---|---|
| 27/07/07 | 27/07/05 | 27/07/03 |
| 27/07/07 | 27/07/07 | 27/07/07 |

Figure 1. Image overlapping example

The result of the combination of comparison methods based on character recognition and adaptive pixel-by-pixel comparison of word images is the found correspondence between the words of the reference and test images. Errors in the image segmentation into words, as well as recognition errors, lead to false positives, that is, modifications that do not actually exist. The number of false positives in comparison results can be reduced by analyzing words that are neighbors of uncoordinated words.

The check of concatenation of two words and the division of the word into several parts, which are the result of incorrect segmentation, was carried out. An example of the case, when most of the split word will be matched to the reference word, and the smaller part will be considered inserted word, is shown in Fig.2.

The flowchart of document image comparison process is shown in Fig.3.

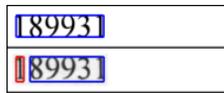

Figure 2. An example of division of a word into two parts. Red marks words that are not coordinated, and blue marks the coordinated words, the top word corresponds to the reference document, and the bottom - to the test

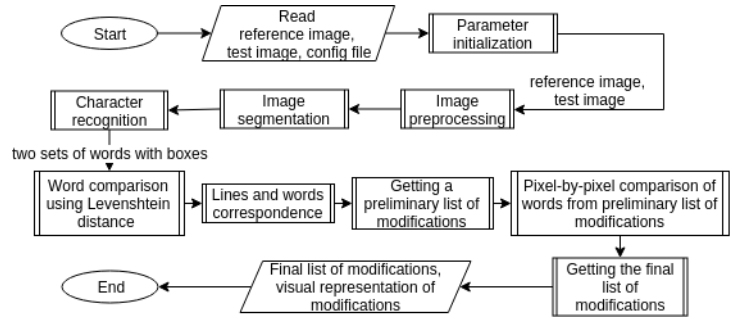

Figure 3. Document image comparison process

## 3. EXPERIMENTS

Public dataset made of a corpus of 477 modified payslips, in which about 6000 characters were modified, was presented in the work [22]. A common problem in the detection of counterfeits is the lack of a public data set to evaluate the algorithms. First, of course, fraudsters do not want to disclose their actions. And second, most of the documents that are subject to modification, contain personal information and are confidential. To solve this problem, authors of [22] synthesized "real" documents that were forged by volunteers who made changes to the document.

The Payslip dataset is divided into three parts, depending on the method of obtaining modifications: intra document copy/paste (CP_Intra), inter document copy/paste (CP_Inter) and Imitation. For the problem that considered in this work, method of obtaining the modification is not significant, so the different ways of obtaining fakes (CP_Inter, CP_Intra, Imitation and Cases 1-3 [22] in the Table 1) will not differ from each other.

The ground truth in this dataset contains information about modified symbols with a bounding box. Since the answer of the described comparison algorithm are modified words, so to estimate the quality of the finding modifications, the characters from the ground truth were combined into words in accordance with the bounding rectangles, and counting modifications occurretabled at the word level. Also, when calculating the results, modifications associated with changing the font of symbols were not taken into account.

The result of the algorithm is a list of modifications with localization in the image that is the coordinates of the bounding box, as well as the number of line in the text and number of word in the line. Also, modifications are shown in the resulting image, an example is presented in Fig. 4. Such an image contains a reference image on the left and the test image on the right. The boxes of words are selected in the image, the blue color is used to select the coordinated words, red - for modified words, and magenta - for deleted and inserted lines.

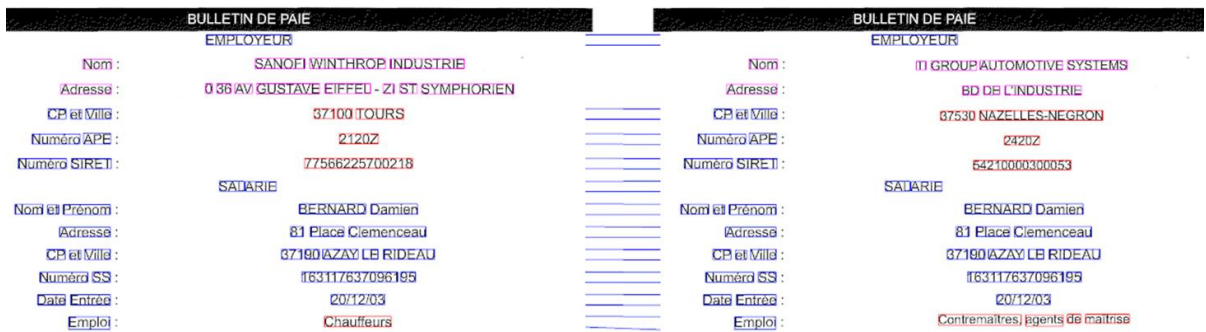

Figure 4. An example of the resulting image with the selected modifications

The first experiment is the comparison of document images, where the comparison of words is carried out using only one method - with the use of character recognition, without using the pixel-by-pixel comparison. As can be seen from the Table 1, the average precision for all pairs of images is 70%, the average recall is 89%, where

$$Precision = \frac{TP}{TP+FP}, \quad Recall = \frac{TP}{TP+FN}. \quad (6)$$

*TP* (true positive) is the number of correctly found modifications. *FP* (false positive) is the number of incorrectly found modifications, that is, the modification is found where it shouldn't have been. And *FN* (false negative) is the number of missed modifications.

The second experiment is the comparison of document images with a combination of the two methods of comparing words: by character recognition and by pixel-by-pixel comparison. The results are presented in the Table 1. The values of $word\_ocr\_simil$ and $word\_pixel\_coeff$ parameters were chosen so that recall values for the two experiments were equal. By changing the value of the parameters it can be achieved higher values of precision or recall. As can be seen, the average precision is 75%, and the average recall is 89%.

|  | CP_Inter | | | CP_Intra | Imitation | | |
|---|---|---|---|---|---|---|---|
|  | Case1 | Case2 | Case3 |  | Case1 | Case2 | Mean |
| Number of pairs | 98 | 58 | 125 | 442 | 28 | 75 | 826 |
| Method without adaptive pixel-by-pixel comparison | | | | | | | |
| Precision | 61% | 83% | 82% | 54% | 81% | 59% | 70% |
| Recall | 84% | 93% | 83% | 91% | 96% | 89% | 89% |
| Method with adaptive pixel-by-pixel comparison | | | | | | | |
| Precision | 71% | 85% | 87% | 61% | 83% | 66% | 75% |
| Recall | 84% | 93% | 83% | 91% | 96% | 89% | 89% |

Table 1. The results of the comparison of image pairs from Payslip dataset with a combination of word comparison methods (CP_Inter, CP_Intra, Imitation and Cases 1-3 denote different ways of obtaining fakes)

Thus, the word comparison method based on the adaptive pixel-by-pixel comparison of word images increased the comparison precision from 70% to 75%. Selecting the parameters that are used in both methods will allow achieving the desired values and the desired balance of precision and recall that are important for specific tasks.

## 4. CONCLUSION

The proposed document image comparison method allows to solve the problem of finding inconsistencies between scanned copies of administrative documents. The combination of character recognition method for words comparison and adaptive pixel-by-pixel method for words image comparison provides high precision and recall of finding modifications. The proposed method can be configured to achieve the required for a specific task values of precision and recall. We have demonstrated the effectiveness of our method in the task of comparing two versions of business documents, at the same time it can be used in different tasks, for example, in document forgery detection. The experiments were carried on the public Payslip dataset (French), consisting of 826 pairs of payslips.

## ACKNOWLEDGMENTS

This paper is partially supported by Russian Foundation for Basic Research (projects 17-29-03170 and 18-07-01384).